\documentclass{article}
\usepackage{arxiv}
\usepackage{graphicx} 
\usepackage{lastpage}
\usepackage{fancyhdr}

\title{\textbf{Simulation Agent: A Framework for Integrating Simulation and Large Language Models for Enhanced Decision-Making}}
\author{\textbf{Jacob Kleiman, Kevin Frank, Joseph Voyles, Sindy Campagna} \\[2ex]
        \textit{AI Research and Emerging Technology Group, PwC US} \\
        \textit{jacob.e.kleiman@pwc.com, kevin.k.frank@pwc.com}}
\date{}
\begin{document}

\maketitle

\begin{abstract}
\vspace{0.5cm}
Simulations, although powerful in accurately replicating real-world systems, often remain inaccessible to non-technical users due to their complexity. Conversely, large language models (LLMs) provide intuitive, language-based interactions but can lack the structured, causal understanding required to reliably model complex real-world dynamics. We introduce our simulation agent framework, a novel approach that integrates the strengths of both simulation models and LLMs. This framework helps empower users by leveraging the conversational capabilities of LLMs to interact seamlessly with sophisticated simulation systems, while simultaneously utilizing the simulations to ground the LLMs in accurate and structured representations of real-world phenomena. This integrated approach helps provide a robust and generalizable foundation for empirical validation and offers broad applicability across diverse domains.
\\[0.5cm]
\end{abstract}

\noindent{\textbf{INDEX TERMS} – AI, Agent, Simulation Model, Digital Twin, LLM}

\vspace{0.5cm}

\section{Introduction}
Acting as conceptual frameworks or computational systems, simulations are crucial for representing the dynamics of complex environments. By simulating system structures, interactions, and feedback loops, they empower scenario analysis, forecasting, and informed decision-making, serving as practical and detailed virtual representations of real-world systems \cite{macal2009agent}. These tools replicate essential system behaviors, causal relationships, and dynamic feedback, enabling in-depth analysis and exploration within controlled, virtual environments \cite{uhrmacher2024context}. However, these modeling techniques, while powerful, often present usability and accessibility barriers to non-technical users due to their inherent complexity \cite{pettinen2005simulation}. As highlighted in research on user-focused cross-domain testing, challenges related to usability and accessibility can significantly impact the interpretability and practical effectiveness of software systems, especially AI-driven ones \cite{de2025towards}. Consequently, running simulations and exploring various scenarios frequently requires substantial technical expertise and manual intervention. Such complexity hinders users' ability to effectively interpret simulation outputs, limiting the models' accessibility, flexibility, and broader practical application. The absence of intuitive, user-friendly interfaces can further restrict their usefulness, particularly in fast-paced business contexts where decision-makers require rapid, comprehensible insights. Ultimately, even the most accurate simulation models may fail to inform decision-making if its outputs remain inaccessible, difficult to interpret, or opaque to the intended users. 

In parallel, LLMs have been explored as potential simulation models due to their natural language interfaces and impressive generative capabilities \cite{rossetti2024social}. However, relying solely on LLMs as representative models can present some challenges. LLMs often lack a fundamental understanding of real-world physics and structured, dynamic processes \cite{hu2023language}. This deficiency leads to "hallucinations" - fabricated or factually incorrect outputs - when LLMs are prompted to act as world models \cite{xu2024hallucination}. The non-transparent nature of LLM outputs, lacking verifiable data or clear reasoning, further undermines their reliability for critical decision-making in business and other high-stakes domains \cite{lavrinovics2025knowledge} \cite{wang2024factuality}. Trust and accuracy are necessary in these contexts; therefore, standalone LLMs can be insufficient. 

Addressing the limitations of simulations and LLMs as independent tools, we propose our simulation agent framework, an integrated, versatile approach designed to combine their strengths. Our framework leverages simulations to provide accurate system representations grounded in real-world dynamics. Concurrently, it employs AI Agents, LLMs enhanced with additional capabilities such as external data access, tool usage, and contextual reasoning and memory, as intelligent interpreters and facilitators. This significantly enhances the adaptability, flexibility, and usability of the overall system. By granting the AI Agent access to simulation models and embedding contextual knowledge of their construction and operation, our simulation agent framework empowers users to navigate simulations, interpret complex results, and test diverse scenarios through natural language interactions. This integration helps make sure that while simulations maintain precision and fidelity, the AI Agent unlocks their accessibility, making sophisticated system analysis available even to non-technical users. Our framework, therefore, offers a robust and adaptable approach capable of supporting effective and informed decision-making across complex and dynamic environments that generalizes to any type of model and use case. 

\section{Related Works}

Recent research at the intersection of simulation modeling and LLMs reflects a growing interest in bridging natural language interfaces with sophisticated computational environments. Broadly, related efforts can be organized into two themes: (1) enabling AI agents to autonomously make decisions and take actions within simulated environments, and (2) harnessing LLMs to improve the usability and explainability of complex modeling processes. Below, we discuss these areas and highlight the gaps that motivate our framework. 

\subsection{AI-Driven Agents in Simulated Environments}

A key line of research has explored integrating LLMs to create believable, autonomous agents capable of rich interactions within simulated worlds \cite{gao2024large}. For instance, previous work demonstrates AI-driven agents that exhibit convincing human-like behaviors within interactive sandbox environments, complete with memory, reflection, and planning \cite{park2023generative}. These generative agents autonomously engage in social interactions, showing emergent behaviors that mimic real-world complexities. 

Similarly, LLMs have been leveraged to simulate diverse users for product design, systematically uncovering a breadth of user requirements through orchestrated interactions among agents \cite{ataei2024elicitron}. Such an approach captures latent user needs that might otherwise be missed through conventional methods. Related research has further underscored the versatility of LLM-based agents by developing language-instructible embodied agents that operate across diverse 3D environments, grounding language understanding in perception and action \cite{raad2024scaling}. Expanding these ideas to social and organizational modeling, recent studies demonstrate how coupling agent-based simulation with LLMs offers rich, context-aware interactions among simulated agents, capturing nuanced dialogues and decision-making processes \cite{gurcan2024llm}. However, grounding these agents in empirically valid models remains challenging, particularly due to a) risks of hallucinations that compromise model fidelity and b) a lack of real evidence supporting decisions. Collectively, this research highlights how LLM-driven agents serve as adaptable, interactive layers for simulated environments but also emphasizes ongoing challenges in robust grounding and transparency. 

Previous research also has explored using multi-agent systems to interact with simulations for optimizing parameter configurations. For example, Xia et al. (2024) introduced agents with specialized roles - observation, reasoning, decision-making, and summarization - designed to control model execution and optimize outcomes \cite{xia2024llm}. In contrast, our simulation agent framework goes beyond optimization. It supports broader analytical exploration by running multiple scenarios in parallel via API calls, without modifying the simulation’s internal code - a key limitation of prior methods. This makes our framework more flexible and easier to integrate across diverse simulation models. 

\subsection{LLMs for Usability, Explainability, and Model Construction}

As simulation models grow more complex, researchers have investigated how LLMs can enhance model usability and interpretability. Recent studies explore generating natural-language explanations of model behaviors, facilitating insights for decision-makers without deep technical expertise \cite{pang2024generating}. This aligns with the broader need for making simulation results comprehensible to non-technical stakeholders. Additionally, AI agents have been used to streamline specialized simulations by helping technical users with code editing, case configuration, and execution, reducing complexity and improving accessibility \cite{xu2024llm}. However, the system referenced was tied to a specific software, lacked simulation model agnosticism, and was designed solely for technical users. Additional efforts have focused on automating the development of system dynamics models by leveraging LLMs. These approaches extract causal structures from text and translate written descriptions into diagrams and models, significantly reducing manual effort \cite{hosseinichimeh2024text}.

\subsection{Research Gap and Motivation}

Although these studies collectively highlight how LLMs can enhance simulation-based methods through improved accessibility, automation, and the creation of realistic agents, most implementations address only part of the overarching challenge of creating robust, user-friendly, and empirically anchored models. Simulation tools remain challenging for non-experts to both configure and interpret, while LLMs, when used in isolation, lack the structured grounding necessary for accurate, trustworthy decision support. Furthermore, existing works often focus exclusively on either facilitating the model setup or interpreting model results, leaving a gap in providing comprehensive interpretability across both phases. 

In contrast, our simulation agent framework seeks to merge the holistic, system-based accuracy of simulation models with the language-driven accessibility of LLMs. By providing the AI agent with direct access to simulation parameters and results, our framework supports interpretability throughout the modeling lifecycle, both in configuring models and interpreting outcomes. Furthermore, the agent possesses the flexibility to determine the optimal way to respond to user queries, whether through clear textual explanations, visual representations such as graphs, or a combination of both. This adaptability can largely enhance user interaction by tailoring complex simulation outputs into intuitive and easily understandable insights, transforming the user's analytical experience. This dual-phase interpretability and flexible communication capability ensure a transparent, verifiable, and engaging decision-making process that is neither reliant solely on text-based reasoning nor constrained by intricate simulation interfaces. This integrated approach positions our framework to substantially advance the usability of simulations for expert and non-expert users alike, offering a novel approach beneficial to a broad array of business, engineering, and organizational domains. 

\section{Challenges Addressed}

Managing the inputs and outputs of simulation models can be complex, especially for users without technical expertise. Our framework simplifies this process with an intuitive, LLM-powered interface, making configurations easier to handle and outputs more accessible.

\subsection{Input and Configuration Challenges}

Configuring simulation inputs is often a complex task, since it requires extensive documentation to guide users in effective simulation setup \cite{karhela2002software}. These inputs define critical elements such as global model parameters, agent characteristics, and event specifications, all necessary for customizing and controlling model execution. Accurate manipulation of these inputs is essential for defining meaningful scenarios, ensuring that simulations produce relevant and accurate outcomes. From our experience, users, especially those external to the model's development, find it difficult to modify, add, or remove parameters without in-depth knowledge of the model's underlying structure. Abstract parameter values can be particularly unintuitive, hindering user understanding and configuration. Consider a parameter like "digital savviness," rated on a scale of 1 to 10. Without clear, accessible documentation and context, users may struggle to grasp its precise definition and implications within the simulation. This lack of clarity can lead to misconfigurations that distort scenario outcomes, undermining the validity of the model’s insights. Similarly, global integer parameters like "choiceFunction," accepting values such as 1, 2, or 3 to specify different algorithms, can leave users uncertain about the functionality associated with each numerical option. If users misinterpret or incorrectly assign these values, the resulting scenario may fail to reflect the intended conditions, diminishing the model's reliability. These examples underscore the critical need for user-friendly tools that simplify the process of understanding, accurately configuring, and effectively manipulating complex model inputs to enable scenario accuracy and meaningful simulation results. 

\subsection{Output Challenges}

Simulation outputs typically generate high dimensional, multi-file time series datasets that pose significant analytical challenges. Even moderate simulations, such as year-long runs with hourly intervals, produce datasets comprising tens of thousands of data points. This complexity escalates rapidly in agent-based models with multiple agent levels, resulting in exponentially larger datasets that quickly surpass users' analytical capacity \cite{lee2015complexities}. Furthermore, scenario explorations and Monte Carlo analyses multiply data volume and complexity, compounding these challenges. 

Beyond sheer volume, effectively interpreting simulation outputs to identify underlying system behaviors and outcome drivers remains a fundamental hurdle. Users often struggle to link simulation data clearly to underlying model dynamics, especially without extensive knowledge of the model’s internal mechanisms. Meaningful insights thus depend heavily on users' comprehension of system logic and their ability to contextualize data within broader simulation behaviors \cite{zhang2024large}. 

In addressing these challenges, LLMs present powerful opportunities. Through interactive data interpretation and natural-language querying capabilities, LLMs facilitate intuitive and efficient exploration of complex simulation outputs \cite{xia2023towards}. This functionality is central to our framework, bridging the gap between intricate datasets and actionable system behavior insights. 

\section{Proposed Framework}

Our framework has five primary components: Simulation Model, Inputs, Outputs, AI Agent, and User. Figure 1 provides an overview of the architecture and the interactions between these core components.

\subsection{Simulation Model}

Our framework leverages specialized software for developing simulation models that supports discrete-event, system dynamics, and agent-based modeling methodologies. These models form the core of our framework, enabling informed decision-making by accurately representing real-world dynamics and integrating real-world data to deliver actionable insights. The model is exported as a standalone application to facilitate streamlined interaction via a single function call. This standalone deployment enables parallel execution of the simulation through asynchronous calls, which is particularly advantageous for efficient scenario evaluation and comparative analysis. 

\subsection{Inputs}

The input data consists of a collection of files that define the configuration of the simulation model, including parameters, simulation agent characteristics, and scenario setups. These files are modified by the AI agent and subsequently read by the simulation model at startup to initialize the simulation environment. Systematic variation of these input files facilitates scenario exploration, enabling comprehensive testing of diverse conditions. This process operates autonomously without direct user intervention, supporting efficient information management and exchange between the simulation model and the AI agent.

\begin{figure}
\includegraphics[width=1\textwidth]{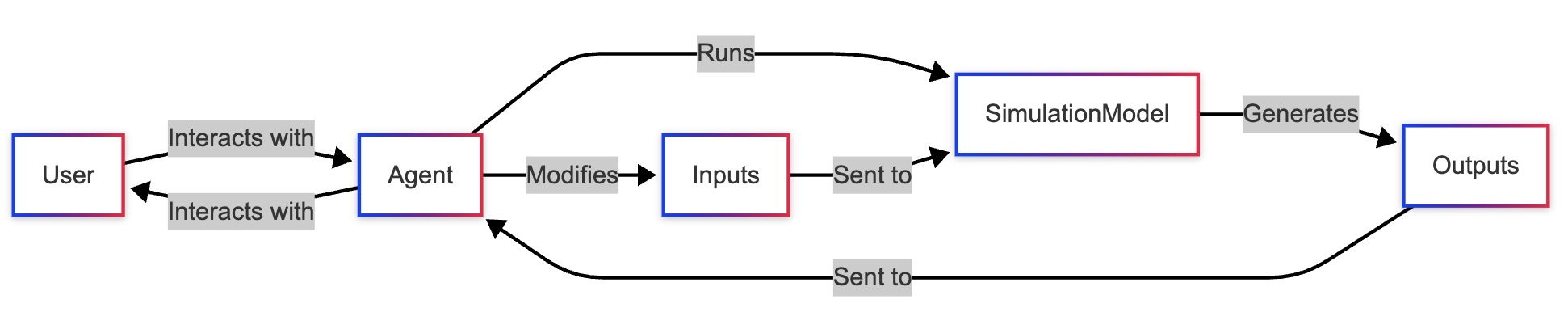}
\caption{\label{Framework} A diagram of the framework, illustrating the interactions between its core components.}
\end{figure}

\subsection{Outputs}

The output data consists of extensive time-series datasets generated by the model. These datasets are structured for efficient post-processing by the AI Agent, enabling dynamic querying of the results based on user-driven questions. This allows the AI agent to provide timely feedback and insights tailored to user needs. The data remains organized and readily retrievable without manual intervention, facilitating automated workflows and deeper analysis of simulation outcomes. 

\subsection{AI Agent}

The AI Agent acts as a crucial bridge between the user and the underlying simulation, inputs, and outputs, significantly reducing complexity and enhancing accessibility for users. Built using the LangChain framework \cite{Chase_LangChain_2022}, the AI Agent integrates an LLM with powerful tool-calling capabilities. This integration enables the AI Agent to perform three key functions: run simulations, modify inputs, and interpret outputs.

\subsubsection{Run Simulations}

The agent is designed to execute simulations autonomously in response to user prompts during a natural language conversation. In our initial implementation, we utilized OpenAI’s GPT-4o as the LLM component \cite{open2023chatgpt}. To facilitate simulation execution, we equipped the agent with a “run simulation” tool, which programmatically initiates the simulation model. 

\subsubsection{Modify Inputs}

A key function of the AI Agent is dynamically modifying simulation inputs. During user interactions, the agent interprets natural language requests to configure the simulation model to set up specific scenarios. The agent translates these instructions into precise modifications of input parameters. To facilitate this, the agent is equipped with a "modify inputs" tool. This allows the agent to update any input parameter by specifying the target file, the relevant field or cell, and the new value. A single user query may require multiple modifications, prompting the agent to make a series of tool calls accordingly. For accurate updates, the agent must understand the purpose of each input file, the role of individual fields, and the meaning of their values. To ensure this, we provide the necessary context in both the tool description fields and the agent’s system prompt.

\subsubsection{Interpret Outputs}

A critical function of the AI Agent is to efficiently process and interpret complex model outputs, especially when dealing with large time-series datasets spread across multiple CSV files. Directly presenting raw data to the agent is ineffective due to its volume and lack of contextual clarity. To address this, the agent uses post-processing tools to query these simulation outputs. One such tool generates JSON summaries highlighting key performance indicators (KPIs), alongside other useful post-processed metrics. These summaries provide the agent with both high-level insights and detailed data necessary to answer a wide range of user queries. 

In our initial experiments, the agent’s system prompt includes detailed information on each data field - its meaning, relevance, and importance - as well as key knowledge about the model’s internal workings, such as critical relationships and causal loops. Typically, only model developers would have access to this level of insight, so exposing it to the agent is transformative. It allows the agent to connect the post-processed data back to the underlying mechanics of the model, providing more actionable interpretations. Ultimately, this deeper context enables the agent to generate insights that go far beyond what a typical user could uncover on their own. 

\subsection{User}

The User is the end stakeholder who interacts with the framework exclusively through natural language queries. Our simulation agent is primarily designed to empower non-technical users, such as business decision-makers, to leverage the power of simulation modeling. However, it has also proven valuable for technical users, including model developers, by streamlining scenario exploration and results analysis. Users of our framework aim to:

\begin{itemize}
    \item Run diverse simulation scenarios to explore different conditions and assumptions.
    \item Understand and compare simulation results across multiple runs to identify key insights and trends.
    \item Ask specific questions about simulation behavior and outcomes, such as "What were the key drivers of these results?" or "How do different levers impact the ultimate KPIs?".
\end{itemize}

Users interact with the AI Agent through natural language to configure these models, execute simulations, and gain actionable insights from the results, all without requiring any direct knowledge of the underlying data files, simulation code, or technical complexities. 

\section{Discussion}

Our simulation agent framework demonstrates a promising approach to helping enhance the usability and accessibility of simulation models by integrating them with LLMs.

\subsection{Usability and Accessibility Gains}

Our framework’s natural language interface significantly lowers the barrier to entry for users interacting with simulation models. Instead of requiring users to navigate complex input files, understand simulation software interfaces, and analyze large datasets or fixed visualizations, our framework allows users to dynamically address their goals and questions with natural language. The AI Agent handles the technical complexities of configuring simulations, executing runs, and interpreting outputs, presenting results and recommendations in an understandable format. This improved usability directly addresses a key challenge in the broader adoption of simulation modeling for decision-making, as highlighted by research emphasizing the importance of accessible and user-friendly simulation tools \cite{uhrmacher2024context} \cite{de2025towards}.

\subsection{Accuracy and Reliability Through Simulation Grounding}

By grounding the model in a robust simulation engine, our framework mitigates the reliability concerns associated with using LLMs as standalone models. While LLMs excel at natural language processing and contextual reasoning, they are prone to hallucinations and lack a fundamental understanding of real-world dynamics \cite{hu2023language} \cite{xu2024hallucination}. Our framework leverages the simulation model to ensure that the insights generated are rooted in verifiable, system-based computations, rather than solely relying on the LLMs potentially flawed internal world knowledge. The LLM acts as an interface and interpreter, not as the source of truth about the system being modeled.

\subsection{Interactive Scenario Exploration}

Our framework enables highly interactive and iterative scenario exploration. Users can easily test "what-if" scenarios by simply describing them in natural language. The AI Agent dynamically translates these requests into simulation input modifications, reruns the simulation, and presents the comparative results. This rapid scenario exploration capability is crucial for effective decision-making in dynamic and uncertain environments, allowing users to quickly assess the potential impacts of different choices and strategies. This interactive capability aligns with the vision of dynamic data-driven simulations that facilitate real-time feedback and optimization \cite{zhang2024large}.

\section{Limitations and Future Directions}

Our simulation agent framework, in its current form, has limitations that point to important directions for future research and development.

\subsection{Enhanced Output Interpretation}

The current approach of equipping the AI Agent with a fixed set of tools for querying output data in predefined ways can be significantly improved by leveraging recent advancements in AI tool-use. Future work should explore more scalable and flexible methods for output data handling and interpretation. Promising directions include representing output data with vector embeddings and utilizing vector databases for semantic querying \cite{kukreja2023vector}, employing advanced LLM-based summarization techniques \cite{ma2023demonstration}, and integrating dynamic tools that enable on-demand code generation and analysis of raw simulation outputs. For instance, emerging systems have demonstrated data science agents capable of autonomously executing end-to-end problem-solving pipelines \cite{hong2024data}.

\subsection{Multi-Agent System}

Another promising direction for future development is transitioning to a multi-agent system architecture. Instead of relying on a single general-purpose agent, this approach would involve a set of specialized agents, each responsible for a distinct aspect of the simulation workflow. For example, one agent could focus on configuring inputs and generating new simulation scenarios, another could specialize in interpreting results and summarizing key findings, and a third could act as an orchestrator to coordinate the workflow across agents. Additionally, a user proxy agent could serve as the primary interface with the end user, translating natural language instructions into structured tasks for the other agents. This modular design could improve scalability, enable more sophisticated capabilities, and allow for clearer separation of concerns across different components of the system.

\subsection{Framework Validation and Domain Generalization}

Our initial experiment focused on evaluating our framework as an AI-powered assistant capable of leveraging simulation models to forecast critical outcomes. To rigorously assess our framework’s effectiveness, quantitative evaluation methods such as measuring prediction accuracy, decision-making support, and user efficiency should be applied to establish a strong performance baseline. 

While early results suggest that the framework is generalizable across various domains and simulation model types, further investigation is needed to validate this. Specifically, additional experiments should be conducted using diverse scenarios, employing consistent and quantitative benchmarks. This approach will help determine the broader applicability and practical impact of the framework across multiple real-world contexts.

\section{Conclusion}

Our simulation agent framework offers a promising approach to modeling by integrating the strengths of simulation models with the strengths of LLMs. By providing a natural language interface to interact with a robust simulation engine, our framework can help lower the barrier to entry for simulation-based analysis, making these powerful tools more accessible to non-technical users. While this paper establishes the foundation for our framework, there remain many opportunities for refinement and expansion. Future research will focus on developing rigorous validation methods to assess the framework’s accuracy, reliability, and scalability across different domains. Additionally, improvements in model integration, adaptability, and computational efficiency will be key areas of exploration. By advancing these aspects, our framework has the potential to broaden the applications of simulation models, empowering a wider range of users to harness its capabilities for analysis, forecasting, and strategic decision-making.

\fancypagestyle{lastpage}{
  \fancyhf{} 
  \fancyfoot[C]{© 2025 PwC US. All rights reserved. PwC US refers to the US group of member firms, and may sometimes refer to the PwC network. Each member firm is a separate legal entity. Please see www.pwc.com/structure for further details. This content is for general purposes only, and should not be used as a substitute for consultation with professional advisors.} 
}

\AtEndDocument{\thispagestyle{lastpage}}

\bibliographystyle{plain}
\bibliography{refs}

\begin{thebibliography}{10}

\bibitem{ataei2024elicitron}
Mohammadmehdi Ataei, Hyunmin Cheong, Daniele Grandi, Ye~Wang, Nigel Morris, and Alexander Tessier.
\newblock Elicitron: An llm agent-based simulation framework for design requirements elicitation.
\newblock {\em arXiv preprint arXiv:2404.16045}, 2024.

\bibitem{Chase_LangChain_2022}
Harrison Chase.
\newblock {LangChain}, October 2022.

\bibitem{de2025towards}
Matheus de~Morais~Le{\c{c}}a and Ronnie de~Souza~Santos.
\newblock Towards user-focused cross-domain testing: Disentangling accessibility, usability, and fairness.
\newblock {\em arXiv e-prints}, pages arXiv--2501, 2025.

\bibitem{gao2024large}
Chen Gao, Xiaochong Lan, Nian Li, Yuan Yuan, Jingtao Ding, Zhilun Zhou, Fengli Xu, and Yong Li.
\newblock Large language models empowered agent-based modeling and simulation: A survey and perspectives.
\newblock {\em Humanities and Social Sciences Communications}, 11(1):1--24, 2024.

\bibitem{gurcan2024llm}
{\"O}nder G{\"u}rcan.
\newblock Llm-augmented agent-based modelling for social simulations: Challenges and opportunities.
\newblock {\em HHAI 2024: Hybrid Human AI Systems for the Social Good}, pages 134--144, 2024.

\bibitem{hong2024data}
Sirui Hong, Yizhang Lin, Bang Liu, Bangbang Liu, Binhao Wu, Ceyao Zhang, Chenxing Wei, Danyang Li, Jiaqi Chen, Jiayi Zhang, et~al.
\newblock Data interpreter: An llm agent for data science.
\newblock {\em arXiv preprint arXiv:2402.18679}, 2024.

\bibitem{hosseinichimeh2024text}
Niyousha Hosseinichimeh, Aritra Majumdar, Ross Williams, and Navid Ghaffarzadegan.
\newblock From text to map: a system dynamics bot for constructing causal loop diagrams.
\newblock {\em System Dynamics Review}, 40(3):e1782, 2024.

\bibitem{hu2023language}
Zhiting Hu and Tianmin Shu.
\newblock Language models, agent models, and world models: The law for machine reasoning and planning.
\newblock {\em arXiv preprint arXiv:2312.05230}, 2023.

\bibitem{karhela2002software}
Tommi Karhela.
\newblock {\em A software architecture for configuration and usage of process simulation models: Software component technology and XML-based approach}.
\newblock VTT Technical Research Centre of Finland, 2002.

\bibitem{kukreja2023vector}
Sanjay Kukreja, Tarun Kumar, Vishal Bharate, Amit Purohit, Abhijit Dasgupta, and Debashis Guha.
\newblock Vector databases and vector embeddings-review.
\newblock In {\em 2023 International Workshop on Artificial Intelligence and Image Processing (IWAIIP)}, pages 231--236. IEEE, 2023.

\bibitem{lavrinovics2025knowledge}
Ernests Lavrinovics, Russa Biswas, Johannes Bjerva, and Katja Hose.
\newblock Knowledge graphs, large language models, and hallucinations: An nlp perspective.
\newblock {\em Journal of Web Semantics}, 85:100844, 2025.

\bibitem{lee2015complexities}
Ju-Sung Lee, Tatiana Filatova, Arika Ligmann-Zielinska, Behrooz Hassani-Mahmooei, Forrest Stonedahl, Iris Lorscheid, Alexey Voinov, J~Gareth Polhill, Zhanli Sun, and Dawn~C Parker.
\newblock The complexities of agent-based modeling output analysis.
\newblock {\em Journal of Artificial Societies and Social Simulation}, 18(4), 2015.

\bibitem{ma2023demonstration}
Pingchuan Ma, Rui Ding, Shuai Wang, Shi Han, and Dongmei Zhang.
\newblock Demonstration of insightpilot: An llm-empowered automated data exploration system.
\newblock {\em arXiv preprint arXiv:2304.00477}, 2023.

\bibitem{macal2009agent}
Charles~M Macal and Michael~J North.
\newblock Agent-based modeling and simulation.
\newblock In {\em Proceedings of the 2009 winter simulation conference (WSC)}, pages 86--98. IEEE, 2009.

\bibitem{open2023chatgpt}
AI~Open.
\newblock Chatgpt (mar 14 version)[large language model], 2023.

\bibitem{pang2024generating}
Andrew Pang, Hyeju Jang, and Shiaofen Fang.
\newblock Generating descriptive explanations of machine learning models using llm.
\newblock In {\em 2024 IEEE International Conference on Big Data (BigData)}, pages 5369--5374. IEEE, 2024.

\bibitem{park2023generative}
Joon~Sung Park, Joseph O'Brien, Carrie~Jun Cai, Meredith~Ringel Morris, Percy Liang, and Michael~S Bernstein.
\newblock Generative agents: Interactive simulacra of human behavior.
\newblock In {\em Proceedings of the 36th annual acm symposium on user interface software and technology}, pages 1--22, 2023.

\bibitem{pettinen2005simulation}
Antti Pettinen, Tommi Aho, Olli-Pekka Smolander, Tiina Manninen, Antti Saarinen, Kaisa-Leena Taattola, Olli Yli-Harja, and Marja-Leena Linne.
\newblock Simulation tools for biochemical networks: evaluation of performance and usability.
\newblock {\em Bioinformatics}, 21(3):357--363, 2005.

\bibitem{raad2024scaling}
Maria~Abi Raad, Arun Ahuja, Catarina Barros, Frederic Besse, Andrew Bolt, Adrian Bolton, Bethanie Brownfield, Gavin Buttimore, Max Cant, Sarah Chakera, et~al.
\newblock Scaling instructable agents across many simulated worlds.
\newblock {\em arXiv preprint arXiv:2404.10179}, 2024.

\bibitem{rossetti2024social}
Giulio Rossetti, Massimo Stella, R{\'e}my Cazabet, Katherine Abramski, Erica Cau, Salvatore Citraro, Andrea Failla, Riccardo Improta, Virginia Morini, and Valentina Pansanella.
\newblock Y social: an llm-powered social media digital twin.
\newblock {\em arXiv preprint arXiv:2408.00818}, 2024.

\bibitem{uhrmacher2024context}
Adelinde~M Uhrmacher, Peter Frazier, Reiner H{\"a}hnle, Franziska Kl{\"u}gl, Fabian Lorig, Bertram Lud{\"a}scher, Laura Nenzi, Cristina Ruiz-Martin, Bernhard Rumpe, Claudia Szabo, et~al.
\newblock Context, composition, automation, and communication: The c2ac roadmap for modeling and simulation.
\newblock {\em ACM Transactions on Modeling and Computer Simulation}, 34(4):1--51, 2024.

\bibitem{wang2024factuality}
Yuxia Wang, Minghan Wang, Muhammad~Arslan Manzoor, Fei Liu, Georgi Georgiev, Rocktim~Jyoti Das, and Preslav Nakov.
\newblock Factuality of large language models: A survey.
\newblock {\em arXiv preprint arXiv:2402.02420}, 2024.

\bibitem{xia2024llm}
Yuchen Xia, Daniel Dittler, Nasser Jazdi, Haonan Chen, and Michael Weyrich.
\newblock Llm experiments with simulation: Large language model multi-agent system for simulation model parametrization in digital twins.
\newblock In {\em 2024 IEEE 29th International Conference on Emerging Technologies and Factory Automation (ETFA)}, pages 1--4. IEEE, 2024.

\bibitem{xia2023towards}
Yuchen Xia, Manthan Shenoy, Nasser Jazdi, and Michael Weyrich.
\newblock Towards autonomous system: flexible modular production system enhanced with large language model agents.
\newblock In {\em 2023 IEEE 28th International Conference on Emerging Technologies and Factory Automation (ETFA)}, pages 1--8. IEEE, 2023.

\bibitem{xu2024llm}
Leidong Xu, Danyal Mohaddes, and Yi~Wang.
\newblock Llm agent for fire dynamics simulations.
\newblock {\em arXiv preprint arXiv:2412.17146}, 2024.

\bibitem{xu2024hallucination}
Ziwei Xu, Sanjay Jain, and Mohan Kankanhalli.
\newblock Hallucination is inevitable: An innate limitation of large language models.
\newblock {\em arXiv preprint arXiv:2401.11817}, 2024.

\bibitem{zhang2024large}
Nan Zhang, Christian Vergara-Marcillo, Georgios Diamantopoulos, Jingran Shen, Nikos Tziritas, Rami Bahsoon, and Georgios Theodoropoulos.
\newblock Large language models for explainable decisions in dynamic digital twins.
\newblock {\em arXiv preprint arXiv:2405.14411}, 2024.

\end{thebibliography}

\end{document}